\documentclass{article}

\usepackage{microtype}
\usepackage{graphicx}
\usepackage{subcaption}
\usepackage{booktabs} 

\usepackage{hyperref}
\usepackage{xurl}

\usepackage[accepted]{icml2026}

\makeatletter
\renewcommand{\ICML@appearing}{Accepted at the ICML 2026 Workshop on \textit{AI4Law}, Seoul, South Korea. Copyright 2026 by the author(s).}
\makeatother

\usepackage{amsmath}
\usepackage{amssymb}
\usepackage{mathtools}
\usepackage{amsthm}
 \usepackage{multirow} 

\usepackage[capitalize,noabbrev]{cleveref}

\theoremstyle{plain}

\theoremstyle{definition}

\theoremstyle{remark}

\usepackage[textsize=tiny]{todonotes}
\usepackage{cleveref}

\icmltitlerunning{Towards a Methodology for HumRights Bench}

\begin{document}

\twocolumn[
  \icmltitle{Toward Human Rights Benchmarking for LLMs: A Pilot Methodology}



  \icmlsetsymbol{equal}{*}

  \begin{icmlauthorlist}
    \icmlauthor{Savannah Thais}{equal,yyy}
    \icmlauthor{Wm. Matthew Kennedy}{equal,oii}
    \icmlauthor{Abhigyan Acherjee}{gu}
    \icmlauthor{Matilda Wysocki}{yyy}
    \icmlauthor{Malcolm Langford}{nio}
    \icmlauthor{Caitlin Kraft Buchman}{ngo}
  \end{icmlauthorlist}

  \icmlaffiliation{yyy}{Hunter College, New York, USA}
  \icmlaffiliation{oii}{Oxford Internet Institute, University of Oxford, UK; King's College London, UK}
  \icmlaffiliation{gu}{Georgetown University, USA}
  \icmlaffiliation{nio}{University of Oslo, Norway}
  \icmlaffiliation{ngo}{AI and Equality, Geneva, Switzerland}

  \icmlcorrespondingauthor{HumRightsBench team}{humrightsbench@gmail.com}

  \icmlkeywords{evaluations, legal reasoning, human rights}

  \vskip 0.3in
]



\printAffiliationsAndNotice{\icmlEqualContribution}

\begin{abstract}
Large language models (LLMs) increasingly mediate legal determinations over  what human rights are realized, and how. Yet, no evaluation benchmark exists to assess whether they can reason correctly about human rights law. To this end, we report our efforts to develop a robust and scalable methodology for creating HumRightsBench—the first expert-validated, scenario-based benchmark for evaluating reasoning grounded in the obligation structure of international human rights law. We adapt the IRAC framework for legal reasoning to better suit the unique reasoning patterns of human rights work (substituting P, "proposing remedies," for C, "legal conclusion," yielding IRAP) to structure our evaluation heuristics. We also produce a pilot series of authentic scenarios designed to implicate the many dimensions of real-world human rights issues and annotated by human rights lawyers and professionals across the world. Ultimately, we find that model accuracy scores range considerably across legal reasoning tasks (overall model performance $\in$ 0.339-0.577, task min-max $\in$ 0.025-0.774), which strongly implies that HumRightsBench is a capable instrument for advancing this emerging subfield of AI evaluations science at a critical moment in its evolution.

\end{abstract} 

\section{Introduction}
Large language models are being deployed at scale in decisions that directly determine whether human rights are realized or violated. Hiring algorithms screen candidates; automated systems adjudicate benefits; content moderation tools govern political speech; procurement processes embed AI into public service delivery \cite{10.1017/err.2024.99}. When actors such as governments, international organizations, technology companies, and civil society organizations make these deployment decisions, they can implicate state obligations under international human rights law to ensure that their actions or activities in their jurisdiction do not contribute to rights violations. Moreover, many individuals and organisations are turning to LLMs for legal advice or building legal advice platforms on top of them \cite{10.1145/3706598.3713470, cole2024navigating}. Yet there is currently no rigorous, principled basis for evaluating whether the LLMs they deploy are capable of correctly practicing human rights legal reasoning.

To work towards addressing this gap, we introduce HumRightsBench, the first expert-validated benchmark for evaluating LLM reasoning across the full arc of human rights legal analysis. Building on the IRAP framework adapted from LegalBench \cite{guha2023legalbench}, we decompose human rights reasoning into four structured subtasks—Issue Identification, Rule Recall, Rule Application, and Proposed Remedies—and develop scenario-based prompts grounded in international human rights instruments, authoritative interpretive guidance, and leading jurisprudence. Scenarios and assessment questions are validated by human rights lawyers and practitioners (mean scenario authenticity $\in$ (0.70-1.00), mean overall question accuracy $\in$(0.633-0.913).

Pilot results across three leading frontier LLMs—GPT-5. Claude Opus 4.7 Gemini 3—are striking. Frontier models cluster near 50-57\% overall accuracy on structured reasoning tasks, demonstrate high stochastic variance across repeated runs, and, on closed-form tasks, perform worst on detecting obligation violations—a foundational practice in human rights law. These results confirm both the scientific utility of the benchmark (models differ in detectable, meaningful ways) and the urgency of the problem (current models are not adequate for human rights reasoning tasks).

This paper makes five contributions: (i) the first benchmark grounded in international human rights law, with a pilot covering the right to water; (ii) an IRAP-based methodology adapted to the structure of human rights legal reasoning; (iii) an expert-validated scenario corpus with documented inter-annotator agreement; (iv) and baseline results across leading LLMs. Section 2 provides background on human rights law and the case for domain-specific evaluation. Section 3 describes benchmark design and methodology. Section 4 presents results. Section 5 discusses implications and future work.

\section{Background}
\subsection{What Are Human Rights?}

The modern human rights regime is one of the most well-developed and universally-agreed bodies of law humanity has produced. Emerging from the cataclysm and horrors of the Second World War as an alternative basis for ordering a world centered on individual human dignity instead of the rights of sovereign states, the human rights project marks a decisive swing towards law as an instrument for making the world as it should be, not for maintaining how it has been \cite{koskenniemi2001gentle, Garcia-SalmonesRovira2013-GARTPO-86}. This swing was a long time in coming. Human rights institution-building drew on more than a century of internationalist thought \cite{sluga2013internationalism} that rejected the centrality of states even in efforts to protect against state abuses of sovereign power \cite{pedersen2015guardians}. Even still, despite the adoption of the Universal Declaration of Human Rights \cite{un_udhr_1948} and a raft of international human rights treaties, decades passed before the human rights regime ascended to dominance over a lingering international order grounded in strong conceptions of state sovereignty \cite{moyn2010utopia}. Though now well entrenched in the global order, the regime still faces continual challenges: noncompliant states, persistent material inequalities, and anti-globalist sentiment \cite{Langford2009-LANSRJ, alston2017populist}. Its consolidation remains incomplete and contested.

Human rights are internationally recognized entitlements that are inherent in all persons by virtue of their humanity, irrespective of nationality, status, or circumstance. They are codified primarily in binding international treaties, such as the International Covenant on Civil and Political Rights (ICCPR, \cite{iccpr1966}, the International Covenant on Economic, Social and Cultural Rights (ICESCR)\cite{ICESCR1966}, and related core instruments including CEDAW \cite{UN_CEDAW_1979}, the CRPD \cite{unCRPD2006}, CRC \cite{UNCRC1989}, and CERD \cite{ICERD1965}, among others. Furthermore, these treaties are interpreted through authoritative soft-law instruments issued by UN treaty bodies, such as General Comments or General Recommendations. General Comments are interpretive guidance documents issued by UN treaty monitoring committees that clarify the scope of treaty obligations without themselves being formally binding. For instance, CESCR General Comment No. 15 elaborates on Article 11 of the ICESCR by, among other things, clarifying that the convention’s “use of the word ‘including’” in introducing the catalogue of rights after the right to adequate standard of living “was not intended to be exhaustive”  (\citep{uncescr2002gc15}, E/C.12/2002/11 p1). In addition, individual UN experts with Special Procedure mandates, appointed by the Human Rights Council, produce thematic or country-specific expert reports. In some instances, these experts are mandated to clarify legal norms (e.g., HRC Res. 7/22, 2008), while in other cases  their work plays this role in practice. Universal Periodic Review (UPR) recommendations are peer-review outcomes generated through the Council’s state-to-state review process. Though not formally binding, these instruments are important to legal determinations and are routinely cited in litigation, policy, and corporate due diligence proceedings. Appendix \cref{tab:hr-sources} recapitulates these various sources and authorities that, together, compose the modern human rights regime.

These instruments impose obligations on a defined set of duty-bearers. Historically and primarily, this has meant states. However, some treaties place obligations on individuals (e.g., the Rome Statute of the International Criminal Court) while some soft law frameworks extend the framework to businesses (e.g, the UN Guiding Principles on Business and Human Rights (United Nations Office of the High Commissioner for Human Rights \cite{ungps2011}) extended the framework to require that businesses---including technology companies and software developers---respect human rights throughout their operations and value chains \cite{ruggie2013, ohchr2019btech}. Under this expanding architecture, the universe of actors with responsibilities to realize rights is becoming broader: it includes governments, international organizations, procurement bodies, civil society, and, critically for this work, the private sector actors who develop and deploy AI systems. Nonetheless, it is primarily states in international human rights law who bear formal legal obligations, although this includes duties to ensure that private actors within their control or influence also respect human rights.

\subsection{Human Rights Work: Prescription or Practice?}

Like any other area of law, human rights law is simultaneously prescriptive and operational. At the prescriptive level, it produces binding obligations and interpretive guidance. At the operational level, human rights practice encompasses the work of litigators, treaty body experts, national human rights institutions, civil society monitors, and corporate due diligence practitioners who translate those norms into determinations about specific situations. This dual character is methodologically significant: a benchmark that captures only doctrinal recall—knowing that the ICESCR guarantees the right to water—will miss the reasoning work that constitutes the field. Competent human rights reasoning requires identifying which obligation is engaged, applying the relevant standard to a factual scenario, and proposing remedies calibrated to the institutional context. 

\subsection{Legal Benchmarking: State of the Field}

We reason that because the human rights regime is sustained by the core text of its provisions as well as the determined efforts of its practitioners to progressively realize those provisions, the rapid diffusion of AI systems into the AI-based decision making systems affecting human rights exerts consequential influence on the project of human rights in general. So too does the proliferation of LLM-powered legal advisory applications and offerings. These effects must be evaluated. It is this full arc of legal and practical reasoning, not proposition recall alone, that HumRightsBench is designed to measure. In so doing, it contributes to an emerging subfield of AI evaluations for law, social impact, and "precursor" capabilities. These fields have developed quickly, albeit with uneven coverage, validity, and robustness. At the same time, promising approaches have arisen. We review some briefly here; full narrative review in Appendix A.

AI evaluation for law has grown quickly. LegalBench assembles 162 expert-built tasks spanning issue-spotting, rule recall, and rule application \cite{guha2023legalbench}, and subsequent benchmarks extend this to law-exam argumentation \cite{fan2025lexam} and structured, IRAC-decomposed reasoning
over real judicial decisions \cite{yu2025mslr, dai2025laiw, fei2023lawbench, li2024lexeval}. A parallel strand measures legal knowledge and language understanding \cite{chalkidis2022lexglue, zheng2021casehold,
chalkidis2023lexfiles, henderson2022pile}, while domain-specific resources target contracts, statutes, and case law \cite{hendrycks2021cuad, wang2025acord, holzenberger2020sara, xiao2018cail, zhong2020jecqa}. Multilingual corpora and evaluation suites have begun to correct the field's
English-centric origins \cite{niklaus2024multilegalpile, niklaus2023lextreme, rasiah2023scale}. A consistent finding is that models handle legal knowledge far more reliably than legal inference, with accuracy collapsing on multi-step reasoning and reasoning models sometimes underperforming despite ``thinking longer'' \cite{fan2025lexam, yu2025mslr, zhang2025thinking}.

Coverage of human rights and international law remains comparatively thin \cite{10.1163/15718107-bja10081}. Early work predicted ECtHR Article violations from case facts \cite{aletras2016predicting, chalkidis2019neural}; more recent benchmarks classify vulnerability in ECtHR decisions \cite{xu2023vechr} and probe how models navigate trade-offs among Universal Declaration rights \cite{samway-etal-2026-language}, with further proposals targeting hard-to-reach populations \cite{unbench2026} and Geneva Convention protections (Kennedy \& Heath 2026). A large adjacent literature on normative, moral, and ethical reasoning \cite{hendrycks2021ethics,
lourie2021scruples, emelin2021moral, ziems2022moral, subramanian2026promoral, jiao2025llmethics} tests value-laden judgment, but grounds it in aggregated preference rather than legal obligation. A final strand asks how structured
legal outputs should be scored, developing LLM-as-judge and rubric-based pipelines \cite{enguehard2025lemajlegalllmasajudgebridging,
shi2026plawbenchrubricbasedbenchmarkevaluating, 10.1007/s10115-026-02703-7}. Across this landscape, no benchmark evaluates reasoning grounded in the obligation structure of international human rights law---the gap HumRightsBench addresses.

\subsection{Why AI Evaluations Specific to Human Rights?}

Despite the field’s activity, gaps still remain. First, evaluation of normative reasoning capabilities in public international law (proportionality, treaty interpretation, and the application of soft-law instruments) remains substantially underrepresented compared to those targeting private and domestic law \cite{chlapanis2024larechr}. Second, few psychometric-based approaches have emerged, and almost all benchmarks evaluate single-turn or short-chain reasoning, despite the multi-turn argumentative structure of actual legal practice \cite{yu2025mslr, fan2025lexam}. This adds experimental instability by introducing, on the one hand stochasticity within the evaluation dataset, and, on the other, requires LLM-as-a-judge scoring pipelines, which adds yet more stochasticity, despite documented methodological improvements \cite{enguehard2025lemajlegalllmasajudgebridging, bavaresco2024llms}. Third, evaluations cluster around assessing the quality of legal reasoning at the expense of other equally important targets, such as outcome prediction or even the stability of internal representations of the legally-relevant search space itself. Notably, each of these gaps implicates a particular challenge (Table 3, Appendix A) in human-rights-specific legal benchmarking \cite{ohchr2024genai}.

\section{Introducing HumRights Bench}

The intersection of AI and human rights spans a wide range of considerations, from how human rights practitioners incorporate AI tools into their monitoring, advocacy, and reporting workflows, to the ways AI systems themselves enable or undermine the realization of rights through their deployment in consequential decisions \cite{ohchr2024genai}. However, a comprehensive evaluation framework cannot meaningfully address all of these dimensions at once. Following extensive consultation with human rights stakeholders—including practitioners, legal scholars, and civil society monitors—we scoped HumRightsBench to probe a specific capability: the ability of LLMs and LRMs to recognize rights violations in situated factual scenarios and to connect those violations to the relevant sources of international human rights law. We believe this represents a critical first step in characterizing whether AI models can reason about human rights at all. A model that cannot reliably identify when a right is engaged, or which instrument governs a given obligation, cannot be trusted to support downstream human rights work, nor can its outputs in adjacent high-stakes domains be meaningfully audited for rights compatibility. Establishing this baseline capability also creates the empirical foundation for subsequent work on controlling how models surface, discuss, and incorporate human rights knowledge into their behaviors and generated responses.

In addition to providing critical insight into the intersection of AI and law, this focus situates HumRightsBench within the broader AI safety and alignment research agenda through a distinctive lens. Mainstream alignment work typically grounds model behavior in elicited human preferences, aggregated value judgments, or constitutional principles derived from general ethical commitments \citep{bai2022constitutional, ouyang2022training}. While valuable, these approaches treat normative content as a matter of preference aggregation rather than legal obligation. HumRightsBench instead anchors evaluation in international human rights law—primarily UN treaties, General Comments issued by treaty bodies, and other authoritative instruments (listed in \citep{ohchr_core_instruments})---which carry determinate legal force and interpretive structure independent of any individual or population’s expressed preferences. This distinction matters methodologically: where preference-based alignment asks what models should do according to aggregated human judgment, a law-grounded benchmark asks what models must recognize according to a body of norms that duty-bearers are formally obligated to uphold and customs all parties are expected to adhere to. The two framings are complementary, but the latter has been substantially underdeveloped in AI evaluation infrastructure despite its direct relevance to the legal exposure of actors deploying these systems at scale.

\subsection{Methodology}
We use the IRAP framework---Issue Identification, Rule Recall, Rule Application, and Proposed Remedies---as the structural backbone for probing legal reasoning about human rights. IRAP is a modification of the IRAC methodology (Issue, Rule, Application, Conclusion) long established in legal pedagogy and practice as a canonical decomposition of how lawyers move from facts to legal conclusions \cite{columbia2022irac}. IRAC has already been validated as an evaluation scaffold for LLM legal reasoning in legal benchmarks, where it has proven effective at isolating distinct sub-capabilities rather than collapsing them into a single end-to-end accuracy score \citep{guha2023legalbench, yu2025mslr}. The substitution of Proposed Remedies for Conclusion better reflects the operational character of human rights practice: practitioners rarely produce binary guilt-or-innocence conclusions and instead must identify institutional, legal, and policy responses calibrated to the duty-bearer and the rights-holder affected \cite{ohchr2006hrba}. This structural fit between IRAP and the actual work of human rights monitoring and reporting is what makes the framework appropriate for our setting.  In addition, it is difficult to determine precisely whether a violation has occurred---and with what certainty---without more detailed scenarios, while it is easier to determine most likely remedies for the most likely violations.

Around this reasoning scaffold, HumRightsBench is organized into four interlocking components. The \textbf{taxonomy} characterizes the space of human rights violations the benchmark is designed to probe, decomposing the domain along descriptive and analytical axes. \textbf{Scenarios} are realistic, narrative-form situations that ground the evaluation in concrete factual settings and \textbf{sub-scenarios} narrow each scenario into specific claims, actions, or impacts that engage a particular legal question. Finally, \textbf{IRAP questions} are generated from each sub-scenario across the four reasoning steps, producing multiple-choice and open-ended prompts whose detailed construction we describe in Section \ref{sec:dataset}. 

\begin{figure}[h]
\centering
\includegraphics[width=\columnwidth]{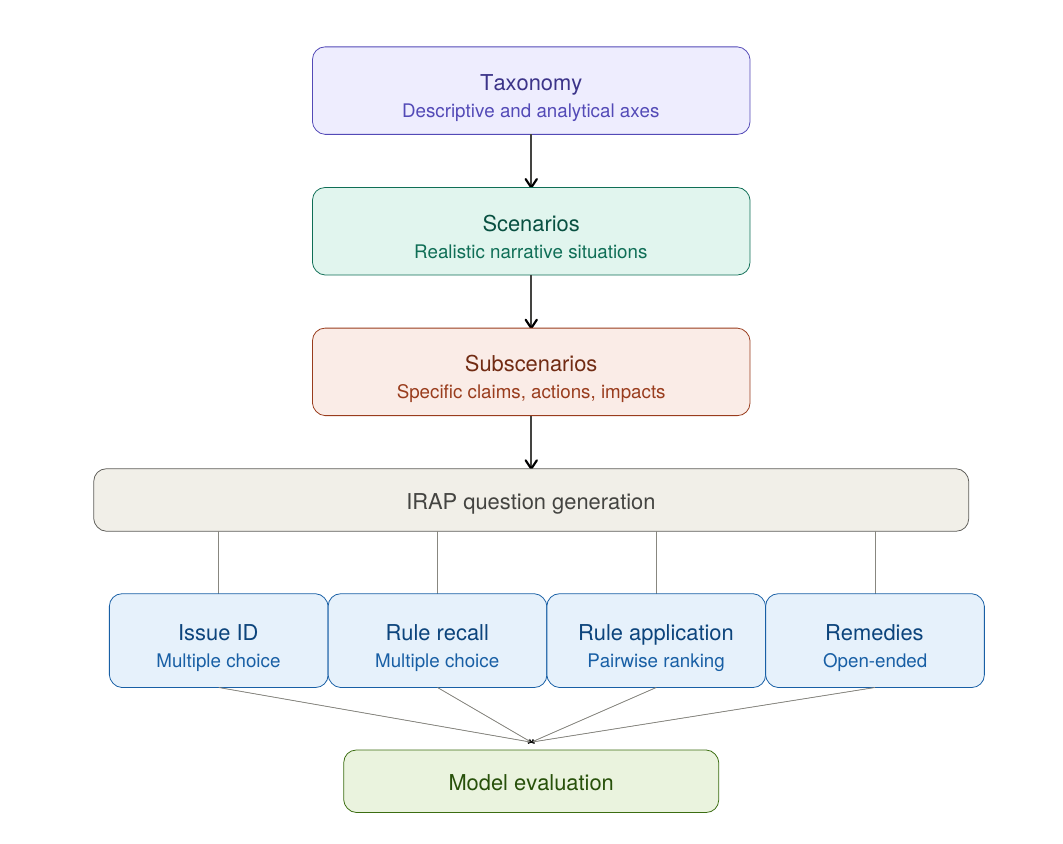}
\caption{Architecture of HumRightsBench. 
}
\label{fig:benchmark-architecture}
\end{figure}

Figure~\ref{fig:benchmark-architecture} illustrates how these components compose into the full benchmark pipeline.

\subsubsection{Taxonomy}\label{sec:taxonomy}
The taxonomy operationalizes the landscape of human rights violations into a structured set of axes against which scenarios are checked for coverage, and supplies the metadata that allows model performance to be decomposed beyond a single aggregate accuracy. We organize it along two families. \textbf{Descriptive axes} characterize the parties involved---who allegedly violated a right and who is affected---without themselves carrying legal valence. \textbf{Analytical axes} characterize the legal structure of the alleged violation: the type of obligation engaged, the character of the failure, and any special situations that modify the standard framework. These axes carry direct legal consequence and are the target of our \textbf{I} questions. 

Tables~\ref{tab:descriptive-axes} and \ref{tab:analytical-axes} enumerate the current taxonomy.
\begin{table}[h]
    \centering
    \small
    \caption{Descriptive axes of the HumRightsBench taxonomy.}
    \label{tab:descriptive-axes}
    \begin{tabular}{p{0.28\columnwidth} p{0.62\columnwidth}}
        \textbf{Axis} & \textbf{Categories} \\
        \hline
        Perpetrator & State; IGO; NGO; company; individual \\
        \addlinespace
        \hline
        Rights-holders & Women; children; people with disabilities; indigenous people; elderly; ethnic minorities; socioeconomically marginalized; migrants; LGBTQIA+; trade unions; political dissenters \\
        \bottomrule
    \end{tabular}
\end{table}

\begin{table}[h]
\centering
\small
\caption{Analytical axes of the HumRightsBench taxonomy.}
\label{tab:analytical-axes}
\begin{tabular}{p{0.28\columnwidth} | p{0.62\columnwidth}}
\toprule
\textbf{Axis} & \textbf{Categories} \\
\midrule
Nature of obligation & Respect; protect; fulfill (conduct and result) \\
\addlinespace
Type of failure & Structural; process; outcome \\
\addlinespace
Type of discrimination & Direct; indirect; intersectional \\
\addlinespace
Special situations & Armed conflict; climate change; AI deployment \\
\bottomrule
\end{tabular}
\end{table}

The descriptive axes draw on the duty-bearer framework articulated in the UNGPs \cite{ungps2011} and the protected groups recognized across the core UN treaties. The analytical axes are grounded in doctrinal scholarship that clusters obligations in different categories:  the respect-protect-fulfill trichotomy (adopted by the UN CESCR (1999)), the conduct-result distinction drawn from the law of state responsibility as developed in the ILC's earlier draft articles (see UN CESCR (1991)), and the structural-process-outcome typology developed in the indicators literature \cite{ohchr2012indicators}. The taxonomy is deliberately layered rather than hierarchical: a single scenario typically engages multiple axes simultaneously, and encoding scenarios against the full set preserves the intersectional character of real human rights situations.

\subsubsection{Scenarios and Sub-scenarios}\label{sec:scenarios}
Scenarios are the narrative anchor of HumRightsBench. Each scenario is a realistic, factually concrete situation that balances the need to achieve authenticity to real-world human rights issues with the need to maintain experimental control via implicating specific elements of the taxonomy described in Section 3.1.1. We deliberately favor narrative scenarios over abstract fact patterns or doctrinal hypotheticals for two reasons. First, the situated factual texture of a scenario—the named setting, the actors involved, the specific resource at stake—is what forces a model to perform rule application rather than rule recitation \cite{guha2023legalbench}, mirroring the analytical work that human rights practitioners do when assessing real situations \cite{ohchr2011manual, aletras2016predicting}. Second, narrative grounding allows us to introduce modular features (geographic location, identity of the affected group, institutional setting) whose values can be varied counterfactually to probe whether model reasoning is stable across protected characteristics---a property that abstract prompts cannot test. Scenarios are drafted from authoritative sources: General Comments, Special Procedures reports, leading jurisprudence, and human rights textbooks. Drafts are reviewed by at least three human rights experts and revised based on annotator feedback before inclusion (Section 3.2). (Section~\ref{sec:annotation}). For the pilot release on the right to water, each scenario is presented to the model as context preceding the IRAP questions as factual substrate against which the questions are answered.

\textbf{Example}: \textit{In the sprawling informal settlement of ``Aqualess Heights'' in the State of Hydronia, thousands of residents, predominantly classified as low-income families, face a daily struggle to access clean and sufficient water. The main water supply consists of a few communal standpipes, often dry or providing water only for limited hours, and privately owned boreholes that charge exorbitant rates equivalent to multiple days of average wages in the area for unsafe water.}

Each scenario is associated with multiple \textbf{sub-scenarios} that narrow the analytical focus to a specific claim, action, or impact. Where the scenario establishes the broad factual setting, the sub-scenario fixes the legal question: it identifies a particular act or omission, attributes it to a specific duty-bearer, and characterizes its effect on identifiable rights-holders. This two-level structure mirrors human rights practice---a country situation is analyzed by isolating discrete events or patterns within it---and it allows the benchmark to generate multiple, semi-independent IRAP question sets from a single scenario without redundant world-building. Two sub-scenarios from the Aqualess Heights scenario above illustrate the range:

\textbf{Sub-scenario A.} \textit{This scarcity forces residents, particularly women and children, to walk for hours to distant, often contaminated, sources, exposing them to health risks like cholera and imposing a significant burden on their time and dignity.} \\
\\
\textbf{Sub-scenario B.} \textit{Despite the national ``Water for All'' policy, there is a chronic under-allocation of public resources to Aqualess Heights. The state's budget priorities have favored large-scale industrial projects over basic service provision in informal settlements, leading to dilapidated infrastructure and insufficient investment in water distribution networks.}

The two sub-scenarios engage different cells of the taxonomy despite sharing a setting: Sub-scenario A foregrounds an outcome failure with intersectional discriminatory impact on women and children, while Sub-scenario B foregrounds a structural failure in the allocation of resources implicating the obligation to fulfill. Each generates its own IRAP question set, and shared world-building across sub-scenarios reduces the annotation burden of scaling the benchmark.

\subsubsection{IRAP Questions}\label{sec:dataset}
Each sub-scenario generates a set of questions structured along the four IRAP steps: Issue Identification, Rule Recall, Rule Application, and Proposed Remedies. The four steps are not interchangeable difficulty levels of the same task; they probe distinct sub-capabilities of legal reasoning, and a model may plausibly succeed at one while failing at another. Separating them allows us to localize where reasoning breaks down rather than collapsing performance into an aggregate score \citep{guha2023legalbench}.

\textbf{Issue Identification (I).} Two multiple-choice questions (I1 and I2) asking (in different ways) which type of failure or obligation violation is most clearly engaged by the sub-scenario, with answer choices drawn from the analytical axes of the taxonomy. As discussed further in Section \ref{sec:annotation}, annotator feedback indicated that some sub-scenarios could plausibly engage multiple failure modes; we revise the question as ``which failure mode is \emph{most} present?'' to elicit the model's primary judgment.

\textbf{Rule Recall (R).} A multiple-choice question presenting a list of legal rules; the model selects the rule that applies. Candidate rules are drawn exclusively from international human rights law and presented uniformly by full instrument name and specific article (e.g., \emph{ICESCR Article 11}) to prevent surface-form cues from substituting for substantive engagement. Each question has only one correct answer. 

\textbf{Rule Application (A).} The model ranks a set of applicable rules by relative authority and relevance to the sub-scenario and provides a short explanation. Our correct answers are constrained by the principle that binding treaty obligations rank above non-binding interpretive instruments. However, as we discuss later, this model only slightly moves the analysis from Rule Recall to Rule Application, and more application focused methodology is discussed in section 5

\textbf{Proposed Remedies (P).} An open-ended short-response question asking the model to propose $<$10 remedies appropriate to the sub-scenario. Remedies must be calibrated to the duty-bearer, the affected rights-holders, and the institutional context of the violation; the open-ended format reflects the irreducibly generative character of remedy proposal.

Together, the four question types trace the full arc of a human rights analysis, from initial issue characterization to identification and application of the governing norms to proposal of responsive measures. Section \ref{sec:annotation} describes how the scenarios, questions, and answers are validated and Section \ref{sec:measurement} describes how each question type is scored.

\subsection{Annotation and Validation}\label{sec:annotation}

HumRightsBench’s validity as a benchmark depends on whether its scenarios are recognizable as authentic human rights situations and are accurate reflections of the field’s understanding of which human rights laws apply. We therefore recruited (see IS1) practicing human rights experts to annotate IRAP questions and judge whether each scenario faithfully serves as a proxy for real-world cases.

We recruited reviewers via posts in four tech policy communities (All Tech is Human, the Center for AI and Digital Policy, Stanford Technology Ethics Program for Practitioners community, and TRUST: The Norwegian Centre for Trustworthy AI). Interested persons were qualified if they had at least two years of experience in human rights legal study or equivalent practice. Ten of 17 applicants recruited through this channel qualified. Six additional human rights experts were recruited through personal networks.

Annotation coverage varied by scenario. Four scenarios received the requisite three raters, and two received two. Three scenarios received no ratings due to high rates of annotator attrition. One scenario received six ratings. Across annotated items, experts broadly agreed that the scenarios represented authentic proxies (\cref{tab:aggregate-rater-results}).

\begin{table}[h]
\caption{Aggregate annotator ratings by question type (5=strongest, 1=weakest; scores of 4 or 5 considered “pass,” all others considered “fail”), and scenario authenticity (higher rate is better). Full IRAP question ratings for each scenario in Appendix D}
\label{tab:aggregate-rater-results}
\vskip 0.15in
\begin{center}
\footnotesize
\setlength{\tabcolsep}{6pt}
\renewcommand{\arraystretch}{1.1}

\begin{tabular}{lc}
\toprule
\textbf{Metric} & \textbf{Mean Score} \\
\midrule

Overall I Questions & 0.729 \\
Overall R Questions & 0.913 \\
Overall A Questions & 0.633 \\
Overall P Questions & 0.850 \\

\midrule

Scenario 1 Authenticity & 0.933 \\
Scenario 2 Authenticity & 1.0 \\
Scenario 3 Authenticity & 0.70 \\
Scenario 4 Authenticity & 0.933 \\
Scenario 5 Authenticity & 1.0 \\
Scenario 9 Authenticity & 0.933 \\

\bottomrule
\end{tabular}

\vskip 0.05in
\parbox{0.95\columnwidth}{\footnotesize
\textit{Note:} No annotations were available for Scenarios 6, 7, and 8.
}

\end{center}
\vskip -0.1in
\end{table}

We were unable to compute IIC (see future work); however, strong ratings coupled with a legible foundational construct (human rights legal reasoning) supports a claim to moderate convergent validity \cite{campbell1959convergent}. Further validation with larger annotator samples is a top priority.

\section{Exploratory Results}

\subsection{Measurement}\label{sec:measurement}

We benchmark four models chosen to cover the current frontier and one open-source reference. An example scenario and subscenario is provided in Appendix~\ref{app:annotation} and the question types are included in Appendix~\ref{app:questions}. The proprietary tier comprises GPT-5 (OpenAI, gpt-5-2025-08-07), Claude Opus 4.7 (Anthropic, claude-opus-4-7-2025-01-30), and Gemini 3 (Google, gemini-3-flash-preview, released 17 December 2025): three flagship systems, included to surface differences across providers at the closed-source frontier. The open-source reference is Qwen~3.5-9B (Alibaba, released 24 February 2026), a 9B-parameter model that fits on a single H100 and approximates what a self-hostable deployment can offer today. Every model is queried with five independent random seeds per question; answer-choice orderings are shuffled per seed to mitigate position bias. All evaluation runs occurred on 20, 21, or 22 May 2026. The overall accuracy across all question types is outlined in Table ~\ref{tab:overall_accuracy}.

\paragraph{Structured-output extraction.} All scoring assumes machine-parseable outputs. For each question type we define a Pydantic schema: a single- or comma-joined answer letter from a multiple choice questions list for I1/I2/R, a comma-separated ranking with per-rule rationale for Ranked Application, and a list of free-text remedies for PR, and obtain conforming responses through each provider's native structured-output interface: OpenAI's \texttt{beta.chat.completions.parse}, Anthropic's tool-use mechanism with the schema declared as the tool input, Gemini's \texttt{response\_json\_schema}, and, for Qwen served via vLLM, JSON-mode generation with the schema injected directly into the prompt. This obviates fragile regex post-processing and ensures grading is performed against exactly the format the model was instructed to produce.

\paragraph{Multiple-choice scoring (I (I1 and I2), R).} The three multiple-choice question types reduce to single- or multi-letter selections. We score each response by exact-set match against the answer key.

\paragraph{Rule Application.} Each Rule Application item, for the purpose of this pilot study, presents a small set of legal rules (typically 5--7) and asks the model to rank them by relevance to a described human-rights scenario. We compute Kendall's $\tau$ (\texttt{scipy.stats.kendalltau}) between the predicted ranking and the gold ranking over the intersection of the two ranked sets, and declare a response \emph{correct} when $\tau \geq 0.7$ --- a conventional cutoff for ``strong'' rank agreement. To avoid spurious credit for models that default to the answer-choice order as presented, the rule labels are shuffled deterministically per (seed, scenario) and the gold ranking is remapped onto the shuffled labels, so the LLM never sees the ground-truth ordering as the natural alphabetical sequence. Responses that fail to produce a parseable, complete ranking are assigned $\tau=0$ and counted as not-correct, matching the errors-as-wrong convention used elsewhere in this section.

\paragraph{Proposed Remedy: embedding-based correctness with human-calibrated thresholding.} To evaluate open-ended LLM responses against reference answers, we score each response by the cosine similarity between OpenAI \texttt{text-embedding-3-small} embeddings of (i) the concatenated ground-truth remedies and (ii) the concatenated model-generated remedies. To convert this continuous score into a binary ``correct''~/~``incorrect'' verdict that can be aggregated across models, we calibrate a single decision threshold $\tau$ against human judgment. Two annotators independently rated $N=40$ responses on a 1--5 Likert scale capturing coverage of the reference remedies; we declare a response \emph{correct} when the mean of the two ratings is at least 4. The threshold $\tau$ is chosen to maximize Cohen's $\kappa$ between the binarized embedding predictor and the human gold label. To avoid optimistic bias from selecting $\tau$ on the same data we evaluate on, we report a leave-one-out cross-validated $\kappa$ in which the threshold is refit on the remaining $n-1$ rows for every held-out example. Inter-annotator agreement (Cohen's $\kappa$ on  binarized labels and Spearman's $\rho$ on raw scores) is reported as a calibration ceiling against which the automatic scorer should be interpreted.

\paragraph{Calibration outcome.} Across $N=40$ annotated responses, inter-annotator agreement was $\kappa=0.02$ on the binarized labels and $\rho=0.42$ on the raw scores. The $\kappa$-optimal threshold was $\tau^{\star}=0.71$, yielding in-sample $\kappa=0.54$ and LOOCV $\kappa=0.12$. We use this $\tau^{\star}$ to report each model's accuracy as the share of responses whose embedding cosine exceeds $\tau^{\star}$, averaged across five seeds per model.

\subsection{Cross-task comparison}\label{sec:results}

\Cref{tab:overall_accuracy} pools all five question types into a single overall accuracy per model. Gemini 3 leads at $0.577 \pm 0.016$, followed by GPT-5 ($0.537 \pm 0.015$) and Claude Opus 4.7 ($0.508 \pm 0.009$); Qwen~3.5-9B lags substantially at $0.339 \pm 0.020$. The per-task breakdown in \Cref{tab:accuracy_by_qtype} sharpens the picture: Claude Opus 4.7 is in fact the strongest model on the R sub-task ($0.774$), but Gemini 3 wins on every other type (I2: $0.710$, RA: $0.240$, PR: $0.630$). Ranked Application is the hardest task across all four models : even the best system clears the $\tau\geq 0.7$ Kendall correctness bar on only roughly one row in four : reflecting the gap between selecting an answer from an enumerated set and producing a fully rationale-aligned ordering. Notably, Qwen~3.5-9B closes most of the gap to the proprietary frontier on PR ($0.531$ vs.\ Gemini's $0.630$) while remaining well behind on the more constrained sub-tasks, suggesting that open-ended generation against a holistic reference is presently more achievable for a 9B open-source model than tasks demanding precise alignment to structured ground truth.
\begin{table}[h]
    \centering
    \caption{Overall mean accuracy across 5 seeds, pooled across I1, I2, R, A, and P question types. RA uses Kendall's $\tau \geq 0.7$; PR uses cosine similarity $\geq 0.71$. Errored LLM calls are counted as incorrect. Qwen 3.5-9B is averaged over 4 seeds (only those with MCQ data present).}
    \label{tab:overall_accuracy}
    \begin{tabular}{lc}
    \toprule
    \textbf{Model} & \textbf{Accuracy} \\
    \midrule
    GPT-5             & 0.537 $\pm$ 0.015 \\
    Claude Opus 4.7   & 0.508 $\pm$ 0.009 \\
    Gemini 3          & \textbf{0.577 $\pm$ 0.016} \\
    Qwen 3.5-9B       & 0.339 $\pm$ 0.020 \\
    \bottomrule
    \end{tabular}
\end{table}
\begin{table}[h]
\centering
\caption{Mean accuracy across 5 seeds, broken down by question type. RA is scored as correct when Kendall's $\tau \geq 0.7$; PR is scored as correct when full-response cosine similarity $\geq 0.71$ (threshold calibrated against human annotators). Qwen 3.5-9B MCQ values (I1, I2, RR) are over 4 seeds.}
\label{tab:accuracy_by_qtype}
\begin{tabular}{llc}
\toprule
\textbf{Model} & \textbf{Question Type} & \textbf{Accuracy} \\
\midrule
GPT-5            & I1 &0.520 \\
GPT-5            & I2 & 0.675 \\
GPT-5            & R & 0.715 \\
GPT-5            & RA & 0.225 \\
GPT-5            & PR & 0.550 \\
\midrule
Claude Opus 4.7  & I1 & 0.473 \\
Claude Opus 4.7  & I2 & 0.595 \\
Claude Opus 4.7  & R & \textbf{0.774} \\
Claude Opus 4.7  & RA & 0.180 \\
Claude Opus 4.7  & PR & 0.520 \\
\midrule
Gemini 3         & I1 & \textbf{0.540} \\
Gemini 3         & I2 & \textbf{0.710} \\
Gemini 3         & R & 0.765 \\
Gemini 3         & RA & \textbf{0.240} \\
Gemini 3         & PR & \textbf{0.630} \\
\midrule
Qwen 3.5-9B      & I1 & 0.394 \\
Qwen 3.5-9B      & I2 & 0.519 \\
Qwen 3.5-9B      & R & 0.494 \\
Qwen 3.5-9B      & RA & 0.025 \\
Qwen 3.5-9B      & PR & 0.531 \\
\bottomrule
\end{tabular}
\end{table}


\section{Discussion and Future Work}

\paragraph{Discussion.} Our pilot results, although limited in scale, suggest that structured human rights reasoning tasks are challenging for frontier LLMs. More interesting is the way in which they are challenging. On closed-form tasks, LLMs regularly performed poorly on issue-identification---a result that has consequential implications not only for human rights workers but also from AI model developers, users, and regulators. Failures at this layer of the human rights reasoning process cascade throughout all other layers, leading to ungrounded rule applications, misconfigured remedies, and, ultimately, produce circumstances in which retrogression becomes structurally more likely. HumRightsBench makes these failures legible to all responsible actors. We note, models perform worst overall on rule application tasks, but this is expected as our pilot methodology sets very high thresholds for ‘correct’ responses here, and we are actively refining these question types. We discuss this further below.

The institutional landscape is now, for the first time, structured to receive this kind of evidence.The Council of Europe's Committee on Artificial Intelligence has adopted HUDERIA as guidance for risk and impact assessment in support of the Framework Convention. HUDERIA is designed to be used by both public and private actors to identify and address risks to human rights, democracy, and the rule of law across all phases of AI deployment. Yet HUDERIA lacks an empirical basis for evaluating whether the LLMs being assessed (or being used to conduct the assessment) are capable of reasoning about the rights implicated. HumRightsBench, once developed, is precisely the tool that can provide such a basis. Similarly, HumRightsBench results could directly inform Fundamental Rights Impact Assessment (FRIA) processes, providing the kind of structured, documented, and reproducible evidence that compliance with Article 27 of the EU AI Act demands.

\textbf{Expanding coverage.} Recall that this pilot's scenarios are limited to a carefully produced selection that implicate only one right primarily: the right to water. Our immediate priority is to achieve more breadth. We plan to extend our scenario coverage to include the \textit{the right to due process} and \textit{the right to education} in the near future. Likewise, we plan to expand into different languages, considering \citep{samway-etal-2026-language}'s demonstration of the variance of model performance across different-language inputs. 

\textbf{Expanding methods.} We will also broaden our task, scoring, and assessment question design. We plan to assess the suitability of implementing an LLM-as-judge scoring pipeline to provide another interpretive signal of model performance on open-ended questions (e.g. the Proposed Remedies). We also plan to further decompose IRAP to include new types of I questions that move away from answer choices predicated on respect-protect-conduct indicators to those that reflect resource-modulated obligations (to minimum core versus progressive realization standards). Additionally, encouraged by promising early signal, we plan to conduct more substantial validation and inter-item consistency testing to ensure the statistical validity of our core construct.

\textbf{Refining assessment.} As noted in Section 3.1.3, our current Rule Application task only slightly moves our analysis from Rule Recall to Rule Application, and, importantly, does not sufficiently demonstrate reasoning over case-specific facts. We are actively developing new types of questions to better assess rule application, namely (1) “rule factors” questions that seek to assess model capabilities to correctly identify appropriate juridical tests, given the facts of specific scenarios; and (2) “jurisprudence” questions, which seek to assess model capabilities to correctly identify appropriate interpretive standards and thereby demonstrate the capability to sample authentic representations of human rights law as it is actually practiced today.

\section{Conclusion}

We introduced the core methodology required to produce HumRightsBench, an expert-validated scenario-based automated evaluation benchmark for human rights legal reasoning in LLMs and LRMs. On a pilot focusing on the human right to water, frontier model performance ranged widely, with significant differences across different elements of our human rights legal reasoning framework (IRAP). Importantly, models performed worst on issue-identification tasks, raising serious questions about their capabilities in this critical area of adoption. Although these results are exploratory--our dataset is small--they establish this area as an urgently important subfield of evaluation science, one that we intend to explore in greater depth and with more robust validation in work to come. 

\section*{Acknowledgments}

We wish to thank all of our scenario reviewers, including Susan Morrissey, Laura Carter, Nathan Heath, Richard Ncube, Selam Abdella, Jera White, Ajitha Sritharan, Angela Kariuki, and others who wish to remain anonymous. We are also grateful to friends and colleagues who advised on several matters throughout the project's inception and delivery, including Dominico Zipoli, Fola Adeleke, Zach Lampell, Helene Moliner, Claudia Flores, Megan Manion, Min Aung, Khalid Hassine, Lynn Gentile, Isabel Ebert, Nathalie Stadelmann, and Jan Rydzak, among others.

\section*{Impact Statement}

\subsection*{IS1. Human Subjects Research for Data Validation} 

Annotation for scenario validation did not require Institutional Review Board (IRB) review, as the activity was determined to constitute PPI rather than human subjects research. Participation was voluntary and consensual: annotators contributed in exchange for acknowledgment rather than compensation, and were informed of these terms at recruitment and again on the survey form itself. Withdrawal was permitted at any time and carried no penalty. No deception was used at any stage. All responses were collected and processed through secure web forms hosted on Google Cloud, accessible only to project staff.

\subsection*{IS2. Ethics Statement}

As we have stated above, we think reporting out methodology positively contributes to the mission shared by many AI evaluation scientists to ensure we develop methods for better understanding the instance- and systems-level impacts and harms AI systems may cause to critical human institutions.

We are mindful, of course, of the dual nature of many technologies. It occurs to us that this tool could in theory instead become a tool for assessing human rights workers themselves. This is not our intent or in the broader human rights community’s interests, and we implore researchers who may build on this work to consider related dual-use risks as AI diffusion and disruption proceeds. We emphatically object to this tool being used to assess the qualifications and performance of human legal professionals, as this would require altogether different methods and artefacts.

Likewise, we reason that such an evaluation could be used by malicious actors as a tool to determine potential loopholes or structural weaknesses in current human rights instruments for the purposes of exploiting them. In future, we intend to carry out precisely this kind of adversarial evaluation to forestall such efforts (and to advance our understanding of model safety guardrail robustness). In any case, we perceive this risk to be low–there is neither sufficient scale nor robustness in this pilot alone to enable such actions–but we note that this is an area of consideration, and we will take measures to prevent such malicious usage as much as possible in future.

\subsection*{IS3. Generative AI Statement}

Authors used generative AI during this research.

In preparation for producing evaluation scenarios, developing a grounding in specific aspects of certain human rights instruments (e.g. social rights conventions) was in part aided by the use of NotebookLM.

Generative AI was also used to aid with table and figure Latex formatting, or, in limited cases, to translate author-written text into an illustrative diagram or summary table.

Likewise, some authors used Generative AI to prepare  bibtex entries from validated sources where no bibtex citation was provided by the publisher.

In all cases, authors retained sole control (and responsibility) for the experimental design, analysis, and interpretation of implications of this work.

\bibliography{example_paper}
\bibliographystyle{icml2026}

\newpage

\onecolumn
\appendix

\section*{Appendix A: Expanded Review of the State of the Art in Legal Benchmarking}
\label{app:litreview}

\subsection*{Summary table of sources of the human rights legal regime}

\begin{table*}[h]
\caption{Sources of international human rights law, sorted by binding force.}
\label{tab:hr-sources}
\caption{Two-tier categorization of international human rights instruments: legally binding treaties versus authoritative but non-binding soft-law instruments, with representative examples for each.}
\centering
\small
\renewcommand{\arraystretch}{1.25}
\begin{tabular}{@{}p{0.15\linewidth} p{0.27\linewidth} p{0.50\linewidth}@{}}
\toprule
\textbf{Category} & \textbf{Instrument Type} & \textbf{Examples} \\
\midrule
\textbf{Binding}
& Core UN human rights treaties (obligations on ratifying states)
& ICCPR, G.A.\ Res.\ 2200A (XXI) (1966); ICESCR, G.A.\ Res.\ 2200A (XXI) (1966); CERD (1965); CEDAW (1979); CRC (1989); CRPD (2006) \\
\midrule
\multirow{6}{=}{\textbf{Authoritative but Non-Binding}}
& Treaty body General Comments / General Recommendations (interpretive guidance from UN monitoring committees)
& CESCR, General Comment No.\ 15: The Right to Water, U.N.\ Doc.\ E/C.12/2002/11 (2002); CCPR, General Comment No.\ 34: Article 19, U.N.\ Doc.\ CCPR/C/GC/34 (2011); CEDAW, General Recommendation No.\ 35 on Gender-Based Violence Against Women, U.N.\ Doc.\ CEDAW/C/GC/35 (2017); CRC, General Comment No.\ 25 on Children's Rights in Relation to the Digital Environment, U.N.\ Doc.\ CRC/C/GC/25 (2021) \\
\cmidrule(l){2-3}
& Special Procedures (thematic and country-specific reports by independent experts appointed by the Human Rights Council)
& Reports of the Special Rapporteur on Extreme Poverty and Human Rights (\cite{alston2015poverty, deschutter2010brazil}); Special Rapporteur on the Right to Food; Working Group on Business and Human Rights; Special Rapporteur on the Situation of Human Rights Defenders \\
\cmidrule(l){2-3}
& Universal Periodic Review (UPR) recommendations (peer-review outcomes from the HRC's state-to-state review cycle)
& UPR Working Group reports issued under HRC Res.\ 5/1 (2007), covering all UN member states on a 4.5-year cycle \\
\bottomrule
\end{tabular}
\end{table*}

\subsection*{Literature Review}
\paragraph{General Legal Reasoning Benchmarks.} The most comprehensive English-language legal benchmark is LegalBench, which assembles 162 tasks across six categories of legal reasoning, including issue spotting, rule recall, and rule application \cite{guha2023legalbench}. LEXam extends this paradigm to long-form legal argumentation, drawing 7,537 questions from 340 law school exams in English and German \cite{fan2025lexam}. Recent Chinese-language work has pushed the field toward structured reasoning evaluation. Like LegalBench, MSLR grounds its tasks in the IRAC framework (a common legal reasoning benchmark decomposing that practice into Issue Identification, Rule Recall, Rule Application, and Conclusion, see section XX.XX below) using real judicial decisions \cite{yu2025mslr}, LAiW organizes evaluation around the legal syllogism in three difficulty tiers \cite{dai2025laiw}, LawBench tests 21 LLMs across 20 tasks spanning memorization, understanding, and application \cite{fei2023lawbench}, and LexEval offers the largest Chinese legal benchmark to date, structured by a cognitive ability taxonomy \cite{li2024lexeval}. Across these benchmarks, model performance degrades sharply when tasks require multi-step reasoning rather than retrieval, indicating that current LLMs handle legal knowledge more effectively than legal inference \cite{fan2025lexam, yu2025mslr}. Counter-intuitively, LRMs occasionally perform much worse than LLMs on certain legal reasoning tasks, despite “thinking longer” \cite{zhang2025thinking}.

\paragraph{Legal Knowledge and Language Understanding.} LexGLUE is the foundational benchmark for legal NLU, modeled on GLUE and covering seven classification and QA tasks across the European Court of Human Rights (ECtHR), the US Supreme Court, EU legislation, and commercial contracts \cite{chalkidis2022lexglue}. CaseHOLD evaluates the identification of holdings in US case law through multiple-choice prompts derived from the Harvard Caselaw Access Project \cite{zheng2021casehold}, while LEDGAR tests multi-label classification of contract provisions from SEC filings \cite{tuggener2020ledgar}. LegalLAMA and LexFiles probe the legal knowledge stored in pretrained models across six common-law jurisdictions \cite{chalkidis2023lexfiles}. Pile of Law provides the underlying 256GB corpus used by many derivative resources \cite{henderson2022pile}. Performance on these benchmarks has saturated relative to specialized legal language models such as Legal-BERT and LegalXLM-R, suggesting that pure language understanding is no longer the binding constraint on legal AI performance \cite{niklaus2023lextreme}.

\paragraph{Contracts, Statutes, and Case Law.} Contract review is the most commercially mature subfield. CUAD contains 13,000 expert annotations across 510 commercial contracts and 41 clause categories, framed as a span-selection task \cite{hendrycks2021cuad}. ACORD extends this to retrieval of precedent clauses for contract drafting \cite{wang2025acord}. For statutory reasoning, SARA tests entailment and question-answering over the US Internal Revenue Code \cite{holzenberger2020sara}, with a follow-up that decomposes statutory reasoning into discrete language understanding challenges \cite{holzenberger2021factoring}. Case-law prediction is dominated by Chinese benchmarks such as CAIL2018 \cite{xiao2018cail} and JEC-QA from the Chinese National Judicial Examination \cite{zhong2020jecqa}. These domain-specific benchmarks consistently show that LLMs perform well on extraction and classification but struggle with tasks requiring the integration of statutory text, case facts, and doctrinal context \cite{holzenberger2020sara, hendrycks2021cuad}.

\paragraph{Multilingual and Cross-Jurisdictional Benchmarks.} As in other fields, the English-only orientation of early legal NLP pretraining corpora has been a persistent limitation \cite{liu2024datasetslargelanguagemodels}. MultiLegalPile addresses this with a 689GB pretraining corpus across 24 languages and 17 jurisdictions, paired with a family of LegalXLM-R models \cite{niklaus2024multilegalpile}. LEXTREME provides the corresponding multilingual evaluation suite, covering classification tasks across European jurisdictions \cite{niklaus2023lextreme}. SCALE tests long-document processing across five languages in the Swiss federal system, with documents extending to 50,000 tokens \cite{rasiah2023scale}. These resources demonstrate that monolingual fine-tuning consistently outperforms multilingual pretraining for jurisdiction-specific tasks, raising open questions about the transferability of legal reasoning across legal systems \cite{niklaus2024multilegalpile}.

\paragraph{Human Rights and International Law.} Human rights NLP began with \citep{aletras2016predicting}, who showed that simple classifiers could predict ECtHR Article violations with 79 percent accuracy from case facts alone---work substantially advanced by \citep{10.1007/s10506-019-09255-y}’s critical engagement. \citep{chalkidis2019neural, chalkidis2022lexglue} subsequently formalized this as the ECtHR-A and ECtHR-B tasks within LexGLUE. VECHR introduces a more demanding objective: classifying vulnerability types in ECtHR decisions and VECHR evaluating model explanations against expert rationales \cite{xu2023vechr}. Most recently, the UDHR Trade-off Benchmark presents 1152 LLM-generated scenarios implicating 24 Universal Declaration articles to test how LLMs reason about trade-offs between rights and competing interests such as public safety or economic stability when using one of eight different evaluation languages \cite{samway-etal-2026-language}. Coverage of substantive international human rights reasoning remains thin relative to domestic legal benchmarks, although this is beginning to change. UNICEF and UNHCR have teamed up to propose a NeurIPS 2026 competition to to validate AI representations of hard-to-reach populations using agency microdata \cite{unbench2026}. More directly, Kennedy and Heath propose a new threat model methodology for evaluating model vulnerabilities to producing deepfake media of POWs violative of several POW protections established in the Geneva Conventions (Kennedy and Heath, forthcoming 2026).

\paragraph{Normative, Moral, and Ethical Reasoning.} Although not legal benchmarks strictly speaking, these kinds of evaluations heavily implicate “precursor” capabilities of interest to legal evaluations and often use legal sources to build evaluation datasets or scoring rubrics. The ETHICS benchmark introduced systematic evaluation of model alignment with datafied representations of justice, deontology, virtue ethics, utilitarianism, and commonsense morality constructs \cite{hendrycks2021ethics}. SCRUPLES extended this to 625,000 ethical judgments over 32,000 real-life anecdotes \cite{lourie2021scruples}, while Social Chemistry 101 cataloged 292,000 rules-of-thumb governing everyday social norms (Forbes et al. 2020). Moral Stories tests norm-consistent and norm-violating action selection in branching narratives \cite{emelin2021moral}, and the Moral Integrity Corpus annotates 38,000 dialogue turns with underlying rules of thumb \cite{ziems2022moral}. More recent work has shifted toward multi-dimensional and adversarial evaluation: MoralBench offers metadata-rich diagnostic structure, ProMoral-Bench unifies evaluation across ETHICS, Scruples, and WildJailbreak under a single moral-safety score \cite{subramanian2026promoral}, and the three-dimensional LLM Ethics Benchmark assesses foundational principles, reasoning robustness, and value consistency \cite{jiao2025llmethics}. A persistent finding across this literature is that models align more closely with individualistic Western moral frameworks than with collectivist or non-Western ones, indicating substantial cultural bias in normative training data \cite{jiao2025llmethics}.

\paragraph{Evaluation Methodology.} A parallel methodological literature has developed around how to score legal and normative outputs. LeMAJ decomposes legal answers into atomic "Legal Data Points" and uses LLM-as-judge scoring validated against LegalBench \cite{enguehard2025lemajlegalllmasajudgebridging}. PLAWBENCH applies rubric-based evaluation specifically to LLM legal agents \cite{shi2026plawbenchrubricbasedbenchmarkevaluating}. LegalEval-Q uses regression-based quality assessment across 49 models to evaluate generated legal text \cite{10.1007/s10115-026-02703-7}. These approaches respond to a recurring concern that simple accuracy metrics inadequately capture the structured, justification-dependent character of legal reasoning \cite{guha2023legalbench, fan2025lexam}.

\clearpage

\section*{Appendix B: Gaps}
\label{app:gaps}

\paragraph{Legal valence.} Human rights norms carry distinctive legal weight that does not obtain in other legal domains. Obligations to respect, protect, and fulfill are legally differentiated, and misidentifying which is at stake produces not merely an inaccurate answer but a legally consequential one. Current safety benchmarks and content policy evaluations are not designed to test this granularity. Existing guardrails---safety classifiers, RLHF alignment, model cards---address human rights concerns only incidentally, through vague values-based framing rather than grounding in the actual normative architecture of international law.

\paragraph{Grounding in practice.} Evaluating models for human rights competencies requires a knowledge of the law and attending normative reasoning practices, but it also requires a scholastic understanding of the application of legal standards to situated facts. That is, it requires an understanding of norms as well as the developed practice of assessing whether certain actions contribute or degrade the progressive realization of those norms. Human rights practice therefore requires accounting for unnamed but specific and heavily implicated duty-bearers, rights-holders, and institutional remedies in concrete scenarios. Instance-level prompting strategies that attempt to elicit human rights reasoning on an ad hoc basis cannot substitute for systematic benchmark-level measurement. Synthetic generation of scenario data instead of expert-creation presents consequential risks to construct validity here \cite{bean2025measuringmattersconstructvalidity, Eriksson2025TrustBenchmarks, nist2024aria_eval_plan, schwartz2025realitychecknewevaluation}.

\paragraph{Highly developed institutional space.} International human rights law has a mature, documented interpretive ecosystem, including treaty bodies, General Comments, Special Procedures, regional courts, and an active body of jurisprudence. This institutional situatedness supplies ample signal towards which to optimize evaluation datasets as well as a demanding standard against which model outputs can be meaningfully assessed. But this requires benchmark specificity–general legal reasoning or legal knowledge benchmarks will not adequately cover this institutional space or its peculiar dynamics. 

\paragraph{Difference from municipal law.} Unlike domestic legal benchmarks, international human rights law is not the law of any single jurisdiction. It operates through treaty ratification, state practice, and authoritative interpretation, without a single apex court (IHL notwithstanding). Reasoning quality cannot be assessed against one jurisdiction's doctrine; it must be evaluated against principles that apply universally while remaining sensitive to implementation variance and the jurisdiction-based legal tests specific judges operating within certain municipal traditions may reach for. This distinguishes it from many existing legal benchmarks, which are grounded in domestic statutory and case law within a single jurisdiction \cite{guha2023legalbench}.

\paragraph{Unique cross-jurisdictional variance.} The progressive realization standard that underpins economic, social and cultural rights in the international human rights regime acknowledges that states implement these rights differently depending on available resources and not only context (which is of course important for all human rights). A benchmark that encodes a single jurisdiction’s approach to a certain case as ground truth without nominating such a context will systematically misrepresent model performance elsewhere. Indeed, this is an instance of a broader problem in hierarchical benchmark design: imposing a uniform taxonomy risks obscuring the context-specific dynamics that determine whether a norm has been violated in practice. Other rubric-based approaches to benchmarking in different fields that apply layered taxonomies representing specific, intermediary, and universal criteria have produced better construct validity \cite{arora2025healthbenchevaluatinglargelanguage}. Exceedingly few have been attempted in legal benchmarking.

\clearpage
\section*{Appendix C: Scenarios}
\label{app:annotation}

\begin{figure}[H]
    \centering
    \includegraphics[width=1\linewidth]{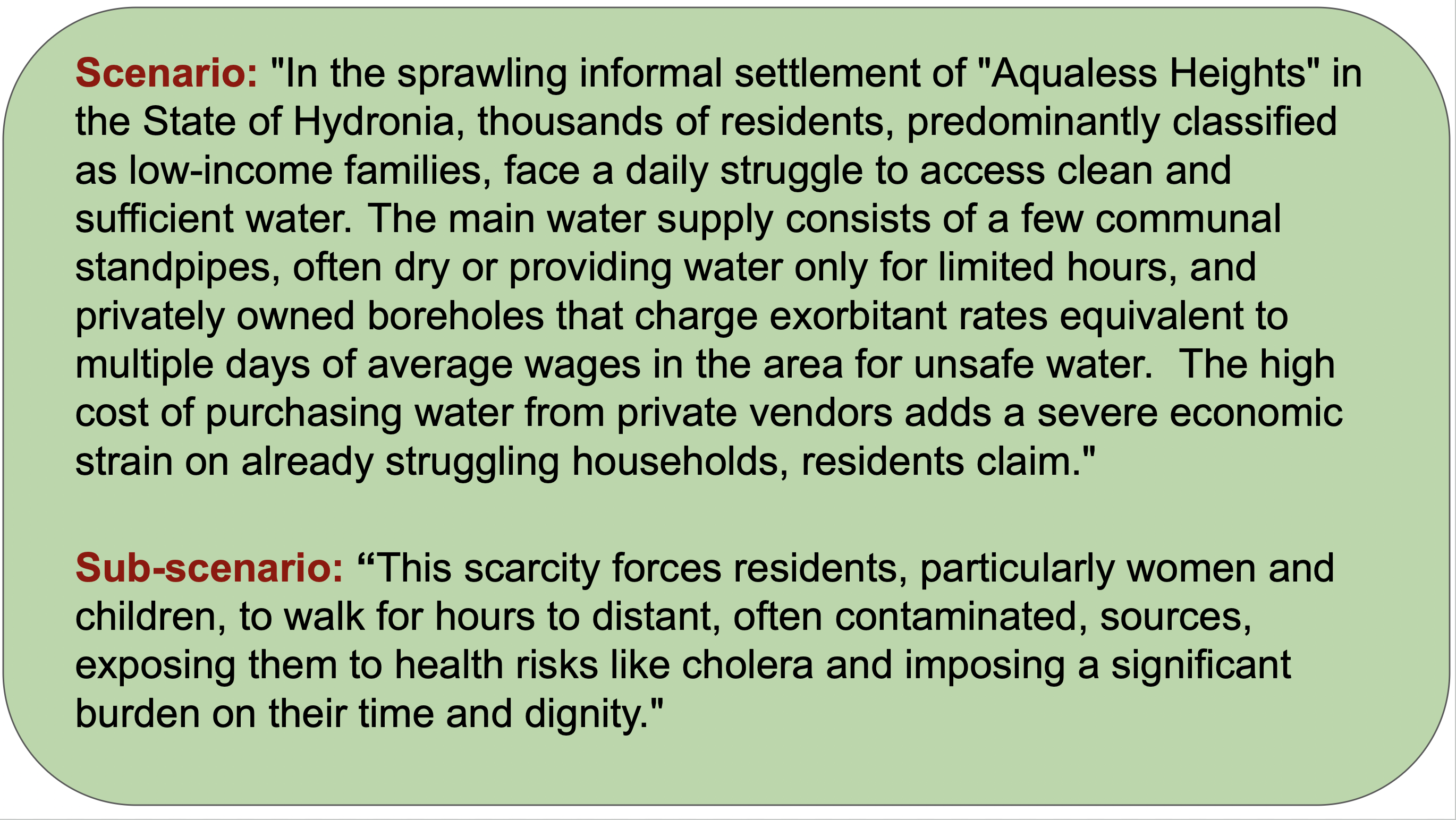}
    \caption{example scenario with sub-scenario}
    \label{fig:scenario}
\end{figure}
\clearpage

\section*{Appendix C: Questions}
\label{app:questions}

There are 5 types of questions in our list.
\begin{figure}[H]
    \centering
    \includegraphics[width=1\linewidth]{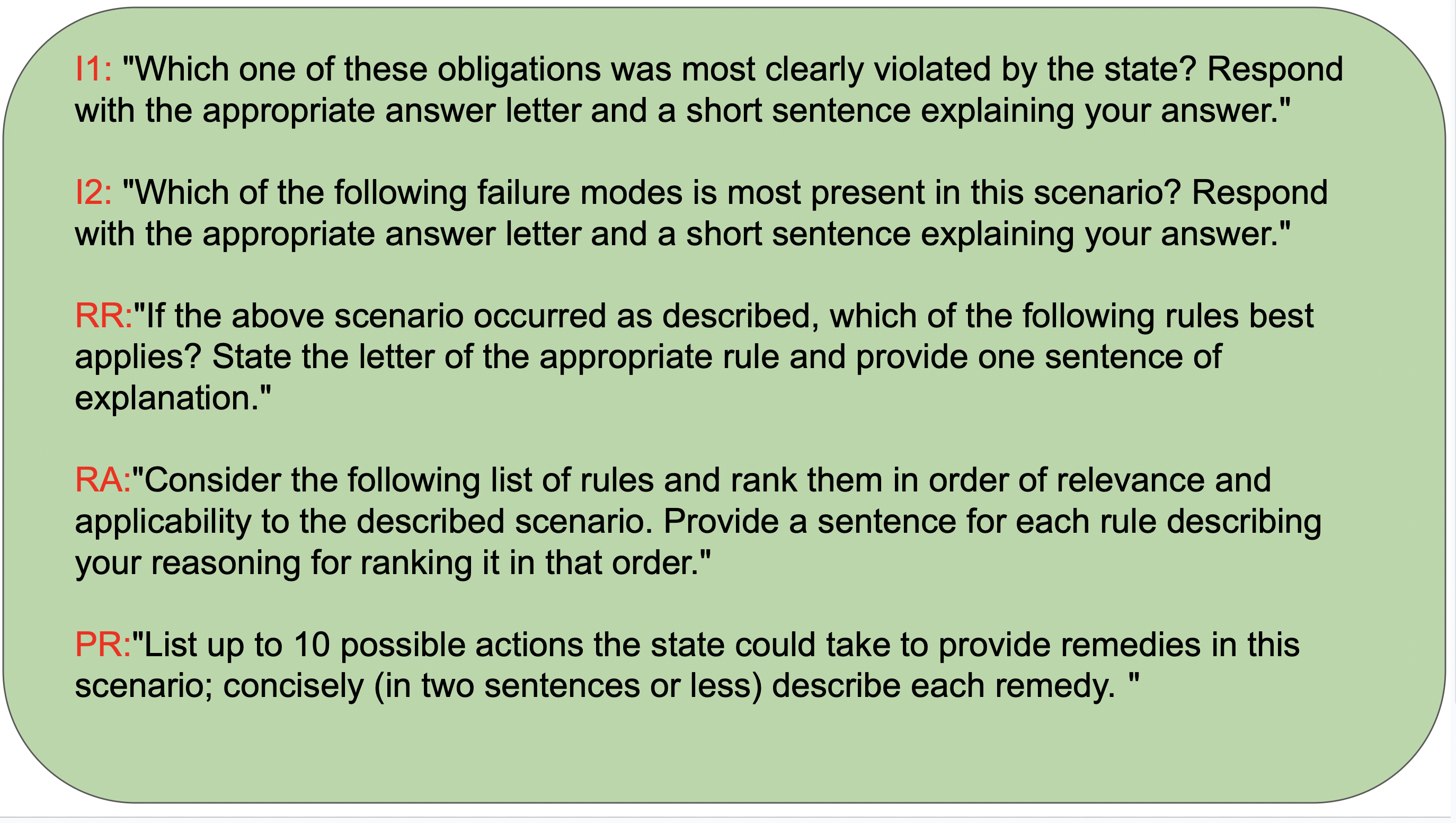}
    \caption{Five questions types in our dataset.}
    \label{fig:placeholder}
\end{figure}

\clearpage
\section*{Appendix D: Annotator ratings of IRAP questions by scenario}

\begin{table*}[h]
\caption{Mean IRAP question ratings by scenario.}
\label{tab:irap-by-scenario}
\vskip 0.15in
\begin{center}
\footnotesize
\setlength{\tabcolsep}{5pt}
\renewcommand{\arraystretch}{1.1}

\begin{tabular}{lcccc}
\toprule
\textbf{Scenario} & \textbf{I} & \textbf{R} & \textbf{A} & \textbf{P} \\
\midrule

Scenario 1 & 0.875 & .75 & 0.50 & 1.00 \\
Scenario 2 & 0.50 & 1.00 & 1.00 & .50 \\
Scenario 3 & 1.00 & 1.00 & 0.50 & 1.00 \\
Scenario 4 & 1.00 & 1.00 & 0.33 & 0.67 \\
Scenario 5 & 0.33 & 1.00 & 0.67 & 1.00 \\
Scenario 9 & 0.67 & 0.73 & 0.80 & 0.93 \\
\textbf{Mean} & \textbf{0.729} & \textbf{0.913} & \textbf{0.633} & \textbf{0.850} \\

\bottomrule
\end{tabular}

\vskip 0.05in
\parbox{0.9\textwidth}{\footnotesize
\textit{Note:} No ratings were available for Scenarios 6, 7, and 8.
}

\end{center}
\vskip -0.1in
\end{table*}

\end{document}